\documentclass{egpubl}
\usepackage{STAG2024}
 
\WsSubmission      

\usepackage[T1]{fontenc}
\usepackage{dfadobe}  

\usepackage{cite}  
\BibtexOrBiblatex
\electronicVersion
\PrintedOrElectronic
\ifpdf \usepackage[pdftex]{graphicx} \pdfcompresslevel=9
\else \usepackage[dvips]{graphicx} \fi

\usepackage{egweblnk} 

\usepackage{caption}
\usepackage{subcaption}
\usepackage{array}
\usepackage{color}
\usepackage{amsmath}
\usepackage{float}

\title[Fast 6D Object Pose Estimation]%
      {FAST GDRNPP: Improving the Speed of State-of-the-Art 6D Object Pose Estimation}

\author[T. Pöllabauer \& A. Pramod \& V. Knauthe \& M. Wahl]
{
\parbox{\textwidth}{
\centering T. Pöllabauer$^{1,2}$\orcid{0000-0003-0075-1181}, A. Pramod$^{1,3}$\orcid{}, V. Knauthe$^{2}$\orcid{}, M. Wahl$^{3}$\orcid{}
}
        \\
{\parbox{\textwidth}{
\centering $^1$Fraunhofer Institute for Computer Graphics Research, Germany\\
         $^2$Technical University Darmstadt, Germany\\
         $^3$University of Siegen, Germany
       }
}
}

\begin{document}

\teaser{
 \includegraphics[width=0.6\linewidth]{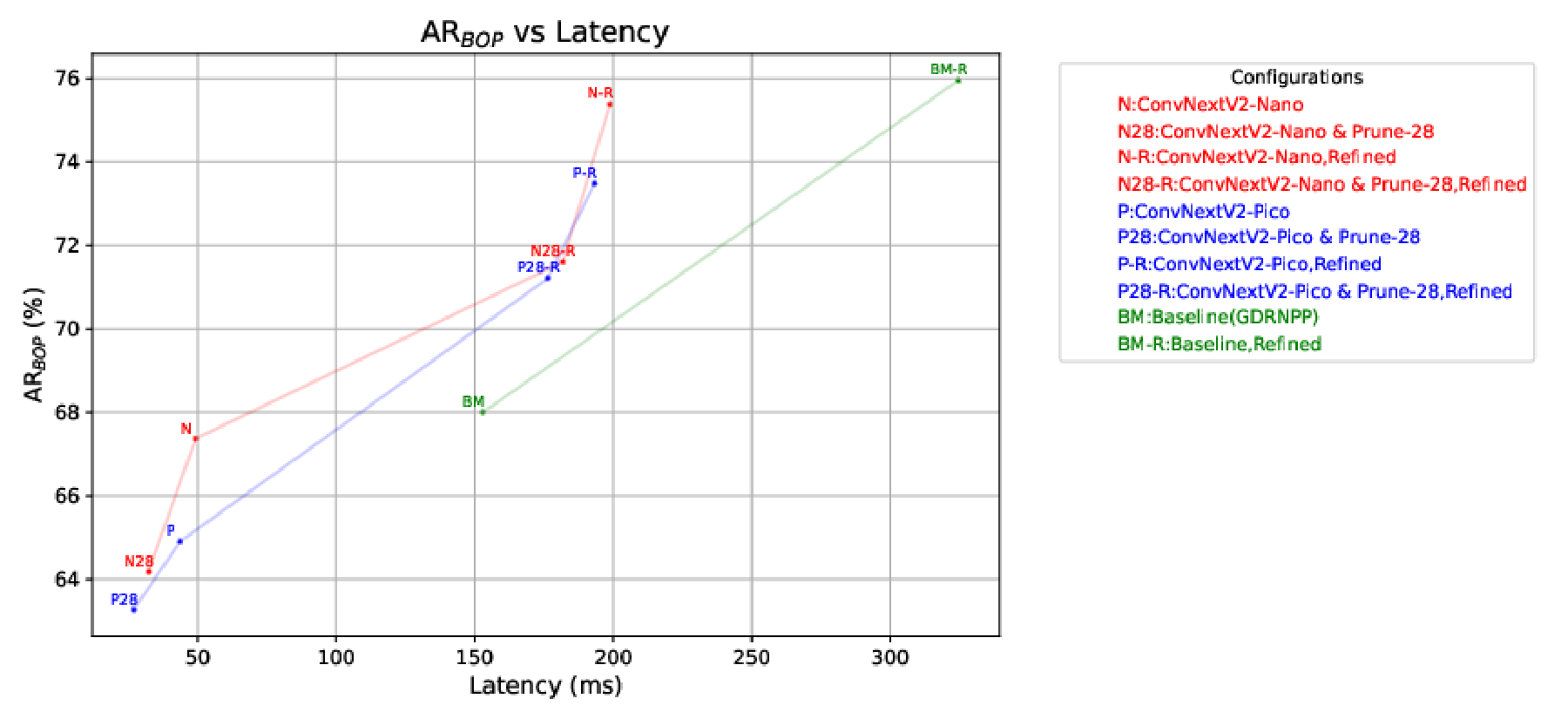}
 \centering
  \caption{Average Recall versus inference speed of our approach against the baseline algorithm GDRNPP evaluated and averaged across all 7 BOP classic core datasets. Our fastest model P28 achieves a 82\% speed-up with only a 4.7\% drop in performance.}
\label{fig:teaser}
}

\maketitle
\begin{abstract}
6D object pose estimation involves determining the three-dimensional translation and rotation of an object within a scene and relative to a chosen coordinate system. This problem is of particular interest for many practical applications in industrial tasks such as quality control, bin picking, and robotic manipulation, where both speed and accuracy are critical for real-world deployment. Current models, both classical and deep-learning-based, often struggle with the trade-off between accuracy and latency.
Our research focuses on enhancing the speed of a prominent state-of-the-art deep learning model, GDRNPP, while keeping its high accuracy. We employ several techniques to reduce the model size and improve inference time. These techniques include using smaller and quicker backbones, pruning unnecessary parameters, and distillation to transfer knowledge from a large, high-performing model to a smaller, more efficient student model.
Our findings demonstrate that the proposed configuration maintains accuracy comparable to the state-of-the-art while significantly improving inference time. This advancement could lead to more efficient and practical applications in various industrial scenarios, thereby enhancing the overall applicability of 6D Object Pose Estimation models in real-world settings. 


\end{abstract}  
\section{Introduction}
Object pose estimation is a fundamental computer vision problem that aims to determine the pose of an object in a given image relative to the camera. Depending on the application needs, the pose can be estimated in varying degrees of freedom (DoF): 3DoF includes only 3D rotation, 6DoF adds 3D translation, and 9DoF further includes the 3D size of the object.
6D object pose estimation, which involves determining both the orientation and translation of an object, is particularly significant in industrial applications. Accurate pose estimation gives an understanding of an object's spatial position, which is vital for tasks in virtual reality, industrial bin-picking, and autonomous driving.
Recent state-of-the-art methods for pose estimation are leveraging multi-stage estimation processes and utilize depth information or scene reconstruction for refinement, leading to good results according to the relevant quality metrics, but inference speed is far from real-time. However, for real-world applications inference time, which must align with real-time data acquisition rates (typically 30 frames per second) is a critical quality for many applications.

Therefore our research focuses on reducing inference time to make pose estimation pipelines deployable in real-time settings and/or on low power devices. Using GDRNPP, an improved version of GDR-Net, as the baseline due to its high performance and modular design, we propose employing model compression techniques to remove dispensable parameters. This compression aims to accelerate the model and reduce its inference time. We evaluate the impact of different parameter reduction methods and their impact on latency versus accuracy across 7 challenging and relevant datasets, taken from the BOP challenge \cite{bop23}.

\section{Related Work}
We will present relevant work on the topic, introducing the BOP challenge and the test datasets (Section \ref{bop}), discussing relevant algorithms (Section \ref{algorithms}) and details on the selected GDRNPP (Section \ref{gdrnpp}), before presenting the ideas of pruning (Section \ref{pruning}) and knowledge distillation (Section \ref{distillation}).

\subsection{BOP Challenge and Datasets}
\label{bop}
With the advent of consumer-grade sensors and cameras, the task of pose estimation has become more accessible. Research and development have progressed in various directions to tackle this complex problem. Meaningful evaluation and comparison of these solutions require unified guidelines. The BOP (Benchmark of Pose Estimation) project, introduced in 2019 \cite{bop20} and repeated in 2022 and 2023 \cite{bop22,bop23}, addresses these needs by providing a comprehensive formulation of the 6D pose estimation task, standardized datasets for performance measurement, a set of pose error metrics accounting for different measurement ambiguities and use case requirements, and an online evaluation system with a leaderboard ranking various state-of-the-art methods. Over the years, new tasks have been added, along with updates to the datasets to better evaluate proposed solutions. Our research focuses on the latest 2023 version of the BOP project. We refer to this iteration to define the 6D pose estimation task on seen objects, compare relevant algorithms, and select the most promising as our candidate for further improvement.

BOP provides a selection of relevant datasets, of which 7 are selected as "core classic". According to the task definition, a algorithm has to be benchmarked on all 7 core classic datasets. A short description of the datasets follows: \textbf{Linemod Occlusion (LM-O).} \cite{brachmann2014learning} An extension of Linemod (LM), designed for evaluating performance in occlusion scenarios. LM contains 15 RGBD sequences with annotated images, CAD models, 2D bounding boxes, and binary masks. It poses challenges with cluttered scenes, texture-less objects, and varying lighting conditions. LM-O includes 1214 RGBD images featuring 8 heavily occluded objects.
\textbf{IC-BIN.} \cite{0be3539b7cdf464eaa71aa69c09ebba3} Focuses on texture-less object pose estimation. IC-BIN addresses clutter and occlusion in robot bin picking scenarios.
\textbf{YCB-Video (YCB-V).} \cite{xiang2018posecnn} Contains 21 objects across 92 RGBD videos, totaling 133,827 frames. It is suitable for object pose estimation and tracking tasks.
\textbf{T-LESS.} \cite{hodan2017tless} Designed for texture-less industrial objects, featuring 30 items from the domain of electrical installation with varying backgrounds, lighting conditions, and occlusions. It is challenging due to the absence of texture and complex object placement.
\textbf{ITODD.} \cite{8265467} Comprises 28 industrial objects across more than 800 scenes with around 3,500 images. It utilizes industrial 3D sensors and high-resolution grayscale cameras for multi-angle observations.
\textbf{TUD-L.} \cite{hodan2018bop} Evaluates the robustness of pose estimation algorithms to lighting variations. TUD-L involves fixed cameras and manually moved objects for realistic movement representation.
\textbf{HB.} \cite{kaskman2019homebreweddb} Covers various scenes with changes in occlusion and lighting conditions, including 33 objects (toys, household items, and industry-related objects) across 13 diverse scenes.


\subsection{Algorithm Candidates for Speed up}
\label{algorithms}

\begin{table}[h!]
\centering
\caption{Results taken from the BOP leaderboard for the 6D localization of seen object tasks \cite{bop_leaderboard}. AR is Average Recall, RGBD uses additional depth input, PBR is using physically-based rendered images for training, Real signals the use of real-world recordings for training, SModel and MModel shows whether a single model per object was trained or whether a single model has to predict multiple objects.}
\begin{tabular}{c}
  \begin{subtable}[t]{\linewidth}
    \centering
    \caption{Performance of the top 10 models, based on the Average Recall score across all seven core BOP datasets, along with the estimation times. Among these methods, only the 10\textsuperscript{th} ranked model uses a single-model approach (SModel), RDPN does not specify which approach they use, and the others employ multi-model approaches (MModel).}
    \label{tab:performance_accuracy}
    \begin{small}
    \begin{tabular}{|c|l|c|c|}
      \hline
        Rank & Method & AR & Time \\
        \hline
        1 & GPose2023 & 85.6 & 2.670 \\
        \hline
        2 & GPose2023-OfficialDetection & 85.1 & 4.575 \\
        \hline
        3 & GPose2023-PBR & 84.4 & 2.686 \\
        \hline
        4 & GDRNPP-PBRReal-RGBD & 83.7 & 6.263 \\
        \hline
        5 & GDRNPP-PBR-RGBD & 82.7 & 6.264 \\
        \hline
        6 & ZebraPose-EffnetB4-refined & 81.3 & 2.577 \\
        \hline
        7 & GDRNPP-PBRReal-RGBD-Fast & 80.5 & 0.228 \\
        \hline
        8 & PFA-Mixpbr-RGBD & 80.0 & 1.193 \\
        \hline
        9 & RDPN & 79.8 & 2.429 \\
        \hline
        10 & GDRNPP-PBRReal-RGBD-OfficialDet. & 79.8 & 6.406 \\
        \hline
    \end{tabular}
    \end{small}
  \end{subtable} \\
  \begin{subtable}[t]{\linewidth}
    \centering
    \caption{Performance of models with fast inference times and an Average Recall of at least 60\% across the seven core datasets. Among these models, the ones ranked 1st, 11th, and 13th use single-model approaches (SModel), while the others employ multi-model approaches (MModel). }
    \label{tab:performance_fast}
    \begin{small}
    \begin{tabular}{|c|l|c|c|}
      \hline
        Rank & Method & AR & Time \\
        \hline
        1 & MRPE-PBRReal-RGB & 69.4 & 0.100 \\
        \hline
        2 & GDRNPP-PBRReal-RGBD-MModel-Fast & 80.5 & 0.228 \\
        \hline
        3 & GDRNPP-PBRReal-RGB-MModel & 72.8 & 0.229 \\
        \hline
        4 & GPose2023-RGB & 72.9 & 0.243 \\
        \hline
        5 & ZebraPose-EffnetB4 & 74.9 & 0.250 \\
        \hline
        6 & ZebraPoseSAT-EffnetB4 & 72.0 & 0.250 \\
        \hline
        7 & ZebraPoseSAT-EffnetB4 (Detection) & 72.0 & 0.250 \\
        \hline
        8 & ZebraPoseSAT-EffnetB4(PBR,Detection) & 72.0 & 0.250 \\
        \hline
        9 & GDRNPP-PBR-RGB-MModel & 70.2 & 0.284 \\
        \hline
        10 & CosyPose-PRBReal & 63.7 & 0.449 \\
        \hline
        11 & GDRNPP-PBRReal-RGB & 67.8 & 0.466 \\
        \hline
        12 & ZebraPoseSAT-EffnetB4+ICP  & 76.5 & 0.500 \\
        \hline
        13 & GDRNPP-PBRReal-RGBD & 74.8 & 0.556 \\
        \hline
        \end{tabular}
        \end{small}
  \end{subtable}
\end{tabular}
\end{table}

To find a suitable algorithm to accelerate, we take a look at the comprehensive BOP leaderboard. Table \ref{tab:performance_accuracy} shows the top-performing models based on Average Recall (AR) across the seven core datasets of the BOP challenge on the task of localizing seen objects. From top to bottom we find places one to three taken up by GPose, an improved version of GDRNPP. GDRNPP \cite{liu2022gdrnpp_bop} is an improved version of  GDR-Net \cite{wang2021gdr}, which again, is an improved version of earlier algorithm CDPN\cite{9009519}. Different parameterizations of GDRNPP take up the places 4, 5, 7, and 10. ZebraPose \cite{su2022zebrapose} comes in on place 6, PFA \cite{hu2022perspective} at 8\textsuperscript{th} place, and RDPN is the last entry in the top 10 at 9\textsuperscript{th} place. Taking inference speed into account and taking another look at the leaderboard sorted by speed we aggregate a second table. Table \ref{tab:performance_fast} focuses on especially fast models according to the reported average processing time per image. To create the table, we sort by speed and only select models with an AR score greater than 0.60 and ignore any algorithm performing worse. We again find all of the above mentioned algorithms in the first positions, with the single additional entry MRPE. 

Our final list of candidates, therefore, consists of MRPE, GPose, ZebraPose, PFA, and GDRNPP. Looking through the list we find MRPE and GPose unpublished and neither does provide implementation details nor code. Comparing the remaining options, the ZebraPose algorithm \cite{su2022zebrapose} uses binary hierarchical encoding for the vertices of a 3D object. It groups and encodes these vertices, storing the mapping along with the 3D vertices offline. A detector identifies regions of interest in a 2D image, and a fully convolutional neural network (CNN) predicts a multi-layer code. This predicted code is then matched with the stored mapping to produce 2D-3D correspondences. Finally, a Perspective-n-Point (P$n$P) solver is used to estimate the 6D pose. RDPN \cite{RDPN} regresses object coordinates per visible pixel and transforms the coordinates into a residual representation to predict pose. Third candidate PFA \cite{hu2022perspective} estimates an initial pose using a first network and matches this pose with offline-generated templates. The comparison between the retrieved template and the target view in the 2D image is performed by computing the displacement field, which is then transformed into 3D-2D correspondences. The final pose is obtained by solving the P$n$P problem using RANSAC/P$n$P. This method consists of two stages and cannot be trained end-to-end. GDRNPP, to be discussed in detail in Section \ref{gdrnpp}, on the other hand is end-to-end differentiable. It follows a modular design, which allows to improve each part of the estimation pipeline separately. At the same time it is the fastest of the top 10 algorithms (7\textsuperscript{th} in Table \ref{tab:performance_accuracy}) and 2\textsuperscript{nd} in Table \ref{tab:performance_fast}, and the base algorithm for best performing algorithm GPose, which, together with other previous extensions \cite{pollabauer2024extending} indicate a flexible algorithm and a worthwhile target for acceleration. ZebraPose and RDPN, in contrast, are noticeably less accurate (5\textsuperscript{th}-8\textsuperscript{th} place in Table \ref{tab:performance_fast}) or twice (12\textsuperscript{th} place in Table \ref{tab:performance_fast}) / ten-times slower (RDPN on 9\textsuperscript{th} place in Table \ref{tab:performance_accuracy}).

Based on this comparison, we choose GDRNPP as our base algorithm. Among the variants of GDRNPP, we choose to select the single-model per dataset version as a baseline. We argue for doing so by highlighting the intended use case: fast pose estimation, for which the loading of a model per object does not make much sense. We also argue that the results as listed in the BOP leaderboard indicate that improvements to the single-model per dataset should propagate to the multi-model approach. We discuss the details of GDRNPP next.

\subsection{GDRNPP}
\label{gdrnpp}
\begin{figure}[!ht]
    \centering
    \includegraphics[width=\linewidth]{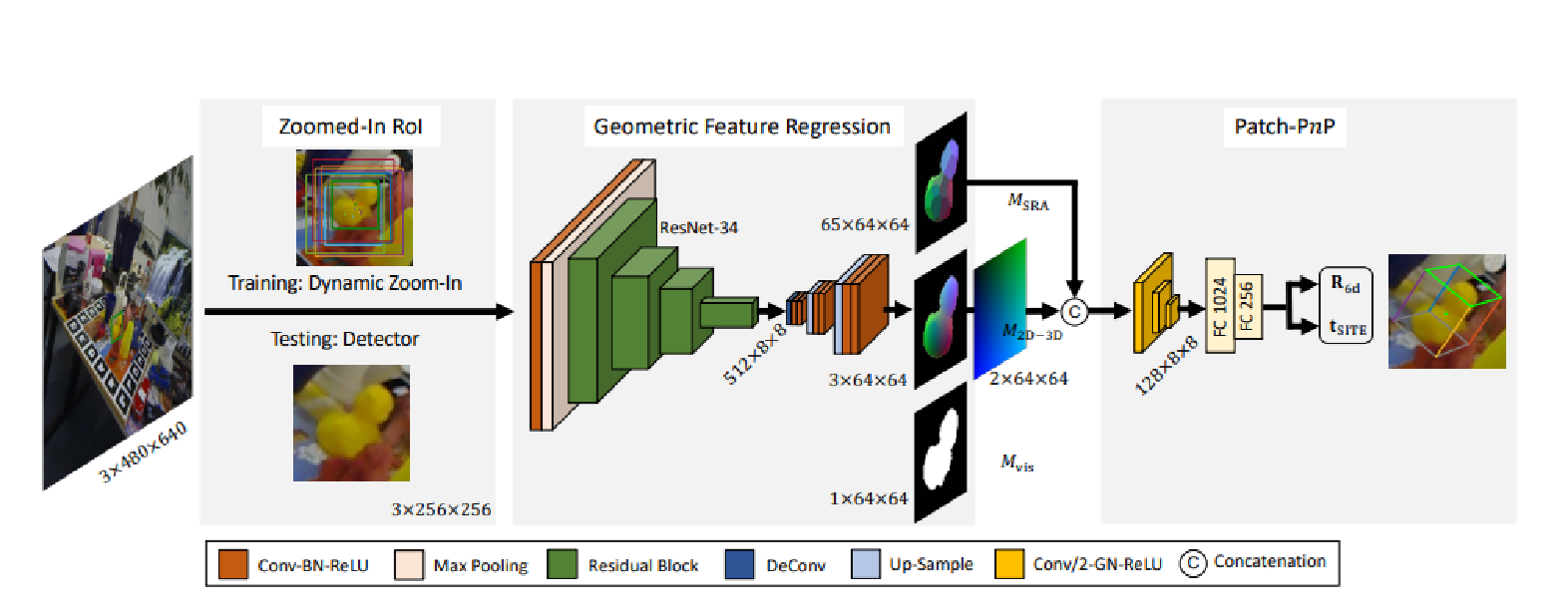}
    \caption{GDR-Net as presented in \cite{wang2021gdr}: Given an RGB image, the model takes a zoomed-in Region of Interest (RoI) of the image as input (1\textsuperscript{st} module). A geometric decoder head then predicts intermediate geometric feature maps (2\textsuperscript{nd} module) $M_{SRA}$, $M_{vis}$, $M_{2D-3D}$. These features are fed into the Patch-P$n$P module (3\textsuperscript{rd} module) to regress the rotation and translation.}
    \label{fig:gdr}
\end{figure}
GDRNPP is an enhanced version of GDR-Net \cite{wang2021gdr} illustrated in Figure \ref{fig:gdr}, using deep neural networks to directly regress required features. It consists of three sub-modules: Object detection, feature prediction, and pose estimation. Also, there is a 4\textsuperscript{th} optional step for pose refinement. To guarantee a fair comparison, we will use the official BOP detections and will not concern ourselves with the detection step at all. The pose refinement module includes two options: Coupled Iterative Refinement \cite{lipson2022coupled} and a fast refinement setup. For our purposes, we will use the fast refinement setup to evaluate the speed improvements introduced by our method. Our primary focus, however, is on both the feature prediction and the pose estimation modules (2\textsuperscript{nd} and 3\textsuperscript{rd} module). 

A forward pass through GDR-Net works as follows (and is functional-wise the same for GDRNPP): An RGB image is fed to the detector, which returns a set of detections. Next, GDR-Net applies dynamic zoom in \cite{Li_2019_ICCV}, to create a range of different object crops, making the algorithm more robust to variances in detection bounding box predictions. The object crops are resized to $256 \times 256$ pixels during the training phase. During testing, it uses inputs from the detector (we use the official YOLOX \cite{DBLP:journals/corr/abs-2107-08430} detections). These inputs are then fed into the geometric feature module, a ResNet-34 \cite{he2016deep} backbone, which extracts image feature maps of resolution $512 \times 8 \times 8$ from the zoomed-in RoI, and a decoder head, which subsequently extracts intermediate geometric features from the image feature maps, namely surface region attention maps $M_{SRA}$, visible/modal object mask $M_{vis}$, and 2D-3D correspondence maps $M_{2D-3D}$. Finally, a Patch P$n$P module directly regresses the rotation (3DoF) and translationb(3DoF) from the learned features $M_{SRA}$ and $M_{2D-3D}$.
The main architectural differences between GDR-Net and GDRNPP are the use of ConvNext-B \cite{liu2022convnet} as a more advanced backbone for feature extraction, the addition of amodal mask prediction $M_{amo}$, in addition to modal mask $M_{vis}$, as well as more involved image augmentation.

\subsection{Pruning}
\label{pruning}
\cite{liang2021pruning,cheng2017survey} survey the use and effects of pruning and quantization. Pruning is the process of reducing the number of learnable parameters in a network by removing those that only negligibly add to the function approximation. Pruning might help to prevent overfitting and improve speed. Quantization is the process of reducing the expressiveness of parameters by switching to lower bit widths, such as 8-bit integers and even smaller. With techniques such as Nvidia's automatic mixed precision \cite{amp} built right into TensorFlow and Pytorch, we focus on pruning. Pruning involves two main steps: first, selecting and removing parameters, and second, retraining the pruned model for a small number of epochs to regain performance, called fine-tuning. Pruning can be done in one-shot, where the network is pruned to the desired degree and then fine-tuned, or iteratively, where the model is partially pruned and retrained multiple times. A further distinction is made between structured \cite{li2022pruning, he2023structured} and unstructured \cite{blalock2020state} pruning. Structured pruning removes entire channels or layers, while unstructured pruning removes individual weights. \cite{vadera2022methods} categorizes common pruning techniques into, first, magnitude based, second, clustering based, and third, sensitivity analysis based methods.


\subsection{Knowledge Distillation}
\label{distillation}

Knowledge distillation (KD) in deep learning describes the process of extracting knowledge from one model (called the teacher) and transferring it to another model (the so called student) \cite{cheng2017survey, gou2021knowledge}. The knowledge transfer can occur from the last layer, the entire teacher model, or specific parts of it, depending on the method used. KD can be applied to arbitrary domains of deep learning, but has been especially interesting for computer vision because of the large networks usually found within the domain. \cite{
wang2021knowledge} provide an in-detail survey on the student-teacher framework applied to computer vision. Also, a very useful property of knowledge distillation, is the fact that models tend to learn faster from a teacher, than from ground truth data \cite{phuong2019towards}.

\section{Methodology}
Next we discuss our approaches to reduce the run-time of GDRNPP. We start introducing a selection of relevant alternative backbones (Section \ref{sec:backboneSelection}), discuss the possibilities of shrinking the decoder head by reducing the number of predicted features (Section \ref{sec:attentionAblation}), applying pruning (Section \ref{sec:pruning}), as well as knowledge distillation (Section \ref{sec:distillation_explanation}).

\subsection{Backbone Selection}
\label{sec:backboneSelection}

GDRNPP originally uses a ConvNext-B model \cite{liu2022convnet} as backbone, featuring 89 million learnable parameters, a top-1 accuracy of 83.8\% on the ImageNet-1K dataset \cite{imagenet_cvpr09}, and 15.6 GFLOPS. For context, the original GDR-Net used ResNet-34 \cite{he2016deep} achieved a top-1 accuracy of 73.31\% with only 3.66 GFLOPS. It is fair to say that ConvNext-B demonstrates better feature extraction capabilities than ResNet-34 and, as a result, much of GDRNPPs outperformance can be credited to the more expressive backbone. The increase in performance between GDR-Net and GDRNPP due to the change in backbone indicates the backbone being a crucial factor for performance. Our goal therefore becomes to find a backbone that performs similar to ConvNext-B, all the while being noticeably faster. 

\begin{table}[h!]
\begin{small}
\begin{tabular}{|c|c|c|c|c|}
\hline
Model & Params & GFLOPs & Throughput & Top-1 \\
\hline
ConvNext-B & 88.6 & 15.4 & 1485 & 83.8 \\
\hline
ConvNext-S & 50.2 & 8.7 & 2008 & 83.1 \\
\hline
FasterVIT-0 & 31.4 & 3.3 & 5802 & 82.1 \\
\hline
FasterVIT-1 & 53.4 & 5.3 & 4188 & 83.2 \\
\hline
FasterVIT-2 & 75.9 & 8.7 & 3161 & 84.2 \\
\hline
FasterVIT-3 & 159.5 & 18.2 & 1780 & 84.9 \\
\hline
ConvNext-V2-T & 28.64 & 4.47 & 1452.72 & 83.8 \\
\hline
ConvNext-V2-N & 15.6 & 2.45 & 2300.18 & 82.1 \\
\hline
ConvNext-V2-P & 9.1 & 1.37 & 3274 & 80.3 \\
\hline
EfficientNet-V2-B0 & 7 & 0.5 & 5739 & 78.7 \\
\hline
EfficientNet-V2-S & 21.5 & 8.0 & 1735 & 83.9 \\
\hline
\end{tabular}
\end{small}
\centering
\caption{Backbone candidates. We identify the following relevant criteria for our selection: number of parameters (millions), GFLOPs, throughput (images per second), and top-1 accuracy (\%).}
\label{tab:backbone}
\end{table}

Table \ref{tab:backbone} lists a selection of, in our eyes, interesting backbone choices, as well as the original ConvNext-B. The data for the first seven models (ConvNext, FasterVIT, and EfficientNet-V2) come from the research by \cite{hatamizadehfastervit}, measured on A100 GPUs with a batch size of 128. The ConvNext-V2 models \cite{woo2023convnext} are evaluated on RTX 3090 GPUs with a batch size of 256. The EfficientNet-V2-B0 is benchmarked on a V100 GPU with a batch size of 128. We based our selection on 4 criteria: total number of learnable parameters, GFLOPS, image throughput, and top-1 classification performance on ImageNet. Since the different models are not benchmarked on the same, though similarly powerful hardware, these numbers need to be taken with a grain of salt, though should give a good estimate for comparisons. To find the best candidates for our pipeline, we will test all of the listed backbones integrated in GDRNPP.

\subsection{Region Ablation}
\label{sec:attentionAblation}
In the paper presenting GDR-Net, the authors test how the number of predicted regions for the attention maps $M_{SRA}$ influence overall accuracy evaluated on LM. They find that already without $M_{SRA}$ (0 regions) performance is good, while increasing the number of regions leads to slight performance gains. This behavior saturates at around 64 regions, which is the final value they settled with, arguing that this is a good trade-off between accuracy and memory requirements. We on the other hand are most interested in the effect on speed, wherefore we re-run the test to answer the question, how different numbers of regions influence run-time. 

\subsection{Pruning}
\label{sec:pruning}

\begin{figure}[!ht]
    \centering
    \includegraphics[width=1\linewidth]{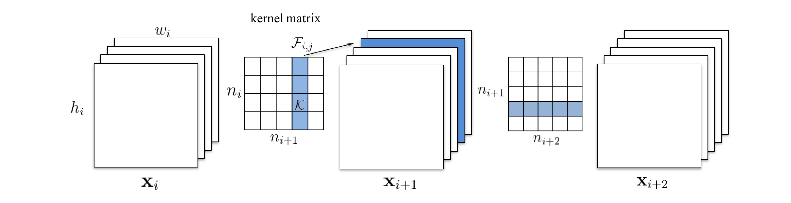}
    \caption{Illustration of \( L_1 \)-norm filter pruning as found in \cite{li2022pruning}. \(F_{i,j}\) is the filter activation on layer \(x_{i}\). Removal of a filter  directly propagates through the network, leading to additional filters being removed further down.}
    \label{fig:prune_filter}
\end{figure}

\begin{figure}[!ht]
    \centering
    \includegraphics[width=1\linewidth]{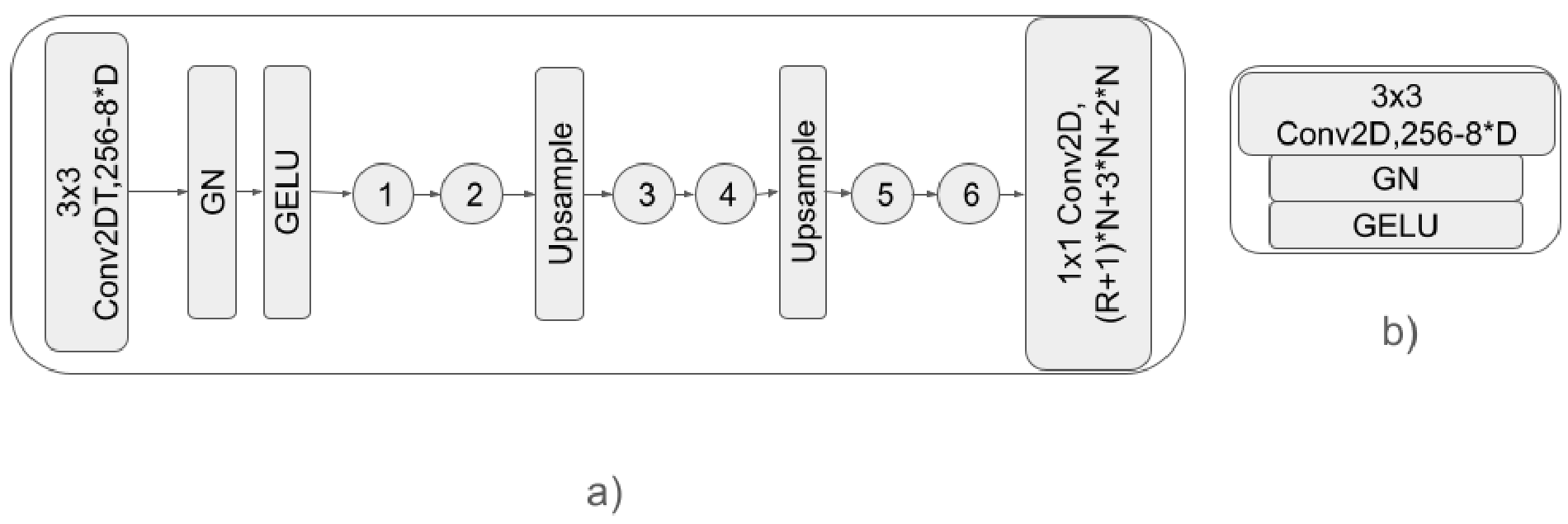}
    \caption{Illustration of \( L_1 \) norm filter pruning applied to the geometric head. The layer architecture is adjusted by \(8 \times D\).}
    \label{fig:prune_ghead}
\end{figure}

\begin{figure}[!ht]
    \centering
    \includegraphics[width=0.8\linewidth]{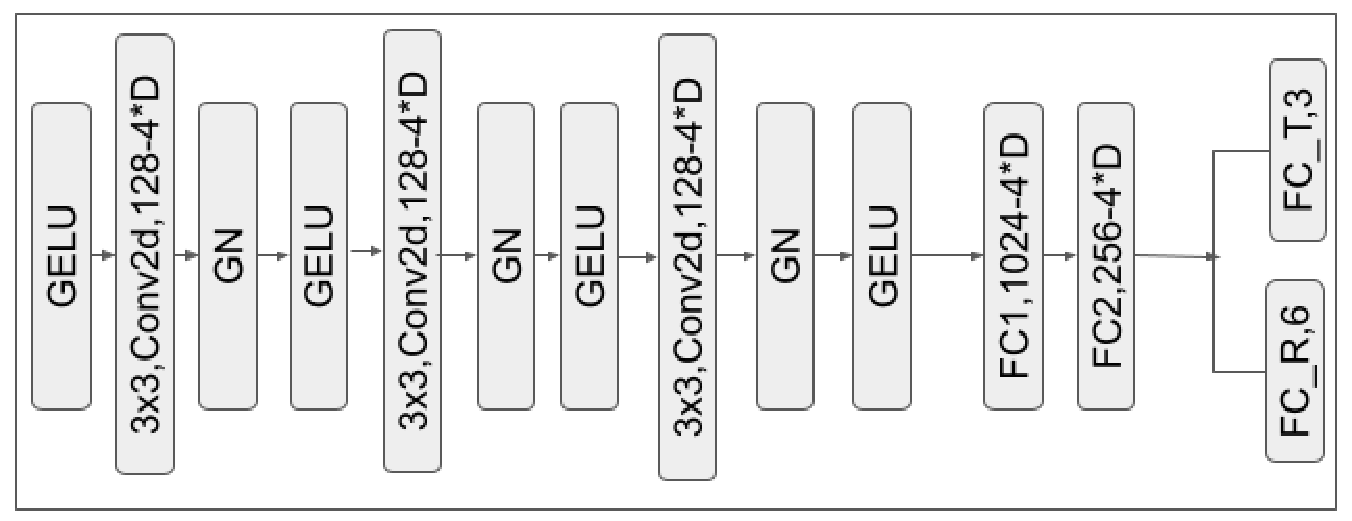}
    \caption{Illustration of \( L_1 \) norm filter pruning applied to the Patch P$n$P module. The layer architecture is adjusted by \(4 \times D\).}
    \label{fig:prune_pnp}
\end{figure}

We apply structured pruning with an \( L_1 \)-norm based filter pruning method \cite{li2022pruning} to the estimation pipeline, removing weights with smaller magnitudes, assuming they have a negligible effect on the model's output. The pruning procedure works as follows: We select the weights of filters in a specific 2D convolutional layer and calculate the \( L_1 \) norm along the channel dimension. Filters are then ranked in ascending order of their norms, and the lowest D-ranking filters are removed. Consequently, the output channels from these removed filters and the kernels applied to these channels are also removed, as illustrated in Figure \ref{fig:prune_filter}.
Given an input feature map \(x_{i}\) with shape \(w_{i} \times h_{i} \times n_{i}\) and a convolutional layer with parameters \(F_{i}\) having shape \(n_{i+1} \times n_{i} \times k \times k\), the total operations are \(n_{i+1} \times n_{i} \times k^{2} \times h_{i} \times w_{i}\). Pruning a kernel \(F_{i,j}\) removes its corresponding feature map, reducing operations by \(n_{i} \times k^{2} \times h_{i} \times w_{i}\). This also affects the next convolutional layer \(F_{i+1}\), further reducing operations by \(n_{i+2} \times k^{2} \times h_{i+2} \times w_{i+2}\). We apply pruning to both the geometric head, as well as the Patch P$n$P module. However, before starting the pruning, the group normalization layers need to be put into consideration, because stochastically selecting and removing layers could disrupt group normalization. Therefore, we remove groups of eight and four filters per parameter D for each module respectively, ensuring the grouping remains intact. Figures \ref{fig:prune_ghead} and \ref{fig:prune_pnp} show the parameterized representation of the layer architecture for both modules with respect to the hyperparameter D, controlling the amount of pruning with a higher value leading to more aggressive pruning.

Once the pruned models are extracted from the larger model following the process detailed above, they are fine-tuned for a few epochs until their performance is close to that of the original, unpruned model.




\subsection{Distillation}
\label{sec:distillation_explanation}
As with any learning problem, the choice of loss function is also very important for knowledge distillation. A typical loss function between the student and teacher is the Kullback-Leibler (KL) divergence loss \cite{hinton2015distilling}, as shown in Equation \ref{eqt:kl}. This loss is calculated between the softened probability distributions of the teacher and student predictions given in Equation \ref{eqt:soft}, controlled by the hyper-parameter $\tau$.

\begin{equation}
L_{KL}(p_s(\tau), p_t(\tau)) = \tau \sum_{j} p_{t,j}(\tau) \log \frac{p_{t,j}(\tau)}{p_{s,j}(\tau)}
\label{eqt:kl}
\end{equation}

\begin{equation}
p_{f,k}(\tau) = \frac{\exp(z_{f,k}/\tau)}{\sum_{j=1}^K \exp(z_{f,j}/\tau)}
\label{eqt:soft}
\end{equation}

Kim et al. \cite{DBLP:journals/corr/abs-2105-08919} compare the effectiveness of KL loss and mean squared error (MSE) loss, finding that MSE performs better for regression-based models. Since our solution space involves regression, we chose MSE as the distillation loss, as shown in Equation \ref{eqt:mse}, where $N$ is the length of the logits vector, $z_{s,i}$ is the student logits, and $z_{t,i}$ is the teacher logits.

\begin{equation}
\text{MSE} = \frac{1}{N} \sum_{i=1}^N (z_{s,i} - z_{t,i})^2
\label{eqt:mse}
\end{equation}

For our distillation process, we select the pre-trained GDRNPP model, specific to each dataset, as the teacher model. Student models are chosen based on the performance of selected backbones, as listed in Section \ref{sec:backboneSelection} and evaluated in Section \ref{sec:backselec}. Given that the output channel dimensions of the student model backbones differ from those of the teacher model, directly applying a loss function without aligning feature dimensions would be problematic. To address this, we initially up-sample the output of the student model's backbone to match the feature dimensions of the teacher model's backbone using a $1x1$ convolutional layer. We then normalize the outputs of both the student and teacher models and apply the MSE loss, as described above. The computed loss is solely used to supervise the student backbone. The geometric head and Patch-P$n$P module are supervised based on their respective loss functions as found in the original GDRNPP \cite{liu2022gdrnpp_bop}. 

\section{Evaluation}
We introduce the task and relevant evaluation metrics (Section \ref{task}), describe the implementation details (Section \ref{sec:implementationDetails}), evaluate the selected backbones integrated in GDRNPP (Section \ref{sec:backselec}), and investigate the influence of reducing the number of regions in $M_{SRA}$ (Section \ref{sec:regions}), the influence of pruning both the geometric head, as well as the Patch P$n$P module (Section \ref{sec:EvalPruning}), and the influence of knowledge distillation (Section \ref{sec:distillation}). Finally we conclude with combining our learnings into our final approach in Section \ref{sec:finalApproach}.

\subsection{Task Description and Metrics}
\label{task}
For a fair comparison we follow the task and evaluation description of the BOP challenge for the task of localization of seen objects. Seen objects refers to the algorithm being allowed to see a set of images of the objects in question during training. The training and test sets of images are disjunct. 
We evaluate the following metrics: latency of the model, Average Recall following the 3 metrics VSD, MSSD, MSPD as defined by BOP, as well as the AR of ADD/ADD-S metric, which is another common metric in the field. The metrics are defined as follows:

\textbf{Latency.}
We measure latency as the forward pass time over all test images per dataset divided by the number of images in the dataset. All measurements, including those for the geometric head and Patch-P$n$P, are taken on the same GPU using the standard time module's performance counter, reported in milliseconds.

\textbf{Average Recall.}
Recall is the ratio of truly positive predictions relative to all positive predictions as defined in Equation \ref{eqt:recall}. 
 \begin{equation}
    \text{Recall} = \frac{\text{True Positives}}{\text{True Positives} + \text{False Negatives}}
    \label{eqt:recall}
    \end{equation}

AR defines thresholds for a given metric. A result is considered correct, if the the prediction falls within the defined threshold. Recall scores are calculated for multiple thresholds and then averaged. For BOP the three recall scores for VSD, MSSD, and MSPD are calculated and averaged to get the final $AR_{BOP}$ score.

\textbf{VSD.}
Equation \ref{eqt:vsd2} calculates VSD by comparing Euclidean distances between distance maps $(\hat{D}, \bar{D})$ rendered from estimated and ground truth poses. The average number of pixels violating the distance threshold within the union of predicted and ground truth masks gives $e_{\text{VSD}}$, with $\tau$ varying from 5\% to 50\% of the object diameter and correctness thresholds ranging from 0.05 to 0.5.
\begin{align}
    e_{\text{VSD}}(\hat{D}, \bar{D}, \hat{V}, \bar{V}, \tau) = \text{avg}_{p \in \hat{V} \cup \bar{V}} \left\{
    \begin{array}{ll}
    0 & \text{if } p \in \hat{V} \cap \bar{V}, \left|\hat{D}(p) - \bar{D}(p)\right| < \tau \\
    1 & \text{otherwise}
    \end{array}
    \right.
    \label{eqt:vsd2}
\end{align}

\textbf{MSSD and MSPD.}

\begin{equation}
    e_{\text{MSSD}} =  \min_{\mathbf{S}\in \mathcal{S}}\max_{\mathbf{x} \in \mathcal{M}} \left\| (\hat{P}\mathbf{x})-(\bar{P} \mathbf{S} \mathbf{x}) \right\|
    \label{eqt:mssd}
\end{equation}

\begin{equation}
    e_{\text{MSPD}} = \min_{\mathbf{S}\in \mathcal{S}} \max_{\mathbf{x} \in \mathcal{M}} \left\| proj(\hat{P}\mathbf{x})-proj(\bar{P} \mathbf{S} \mathbf{x}) \right\|
    \label{eqt:mspd}
\end{equation}

Equations \ref{eqt:mssd} and \ref{eqt:mspd} define MSSD and MSPD respectively. Each vertex in the 3D model $M$ is transformed using the ground truth $\bar{P}$ and estimated pose $\hat{P}$, and the euclidean distances between the transformed 3D points and their 2D projections are computed separately. The minimum distances across all symmetries, $e_{\text{MSSD}}$ and $e_{\text{MSPD}}$, are calculated, with $e_{\text{MSSD}}$ using the same threshold $\tau$ and $e_{\text{MSPD}}$ using a threshold of $5r$ to $50r$ in $5r$ steps, where $r=w/640$ and $w$ is the image width.
    
\textbf{ADD(-S).}

\begin{equation}
    e_\text{ADD} = \frac{1}{m} \sum_{\mathbf{x} \in \mathcal{M}} \left\| (\hat{P}\mathbf{x})-(\bar{P} \mathbf{x}) \right\|
    \label{eqt:add}
\end{equation}

For the ADD metric, the average distance between model points transformed using the estimated and ground truth poses is calculated using Equation \ref{eqt:add}. Recall is computed for correctness thresholds of 2\%, 5\%, and 10\% of the object's diameter and averaged, which can be made symmetry-aware (ADD-S) by measuring the distance to the closest point on the ground truth model.

\subsection{Implementation Details}
\label{sec:implementationDetails}
Our experiments are conducted using a HPC cluster equipped with NVIDIA A100 SXM4 40GB GPUs. We use the BOP toolkit for evaluation, but measure latency ourselves due to instability in the toolkit's latency measurements. Among the 7 core datasets, we select LM-O for our ablation studies because it has fewer training instances but presents challenging scenarios, making it suitable for both computational and benchmarking purposes. We test all latency improvement methods on this dataset to inform our decision process on what to keep for our final model. Based on the performance improvements demonstrated by each method, we generalize to select the configuration that offers the best balance between speed and accuracy discussed in Section \ref{sec:finalApproach}. As for data, we follow the approach of using PBR and REAL data as provided by the BOP challenge.

\subsection{Backbone Selection}
\label{sec:backselec}
\begin{figure}[!ht]
    \centering
    \includegraphics[width=0.8\linewidth]{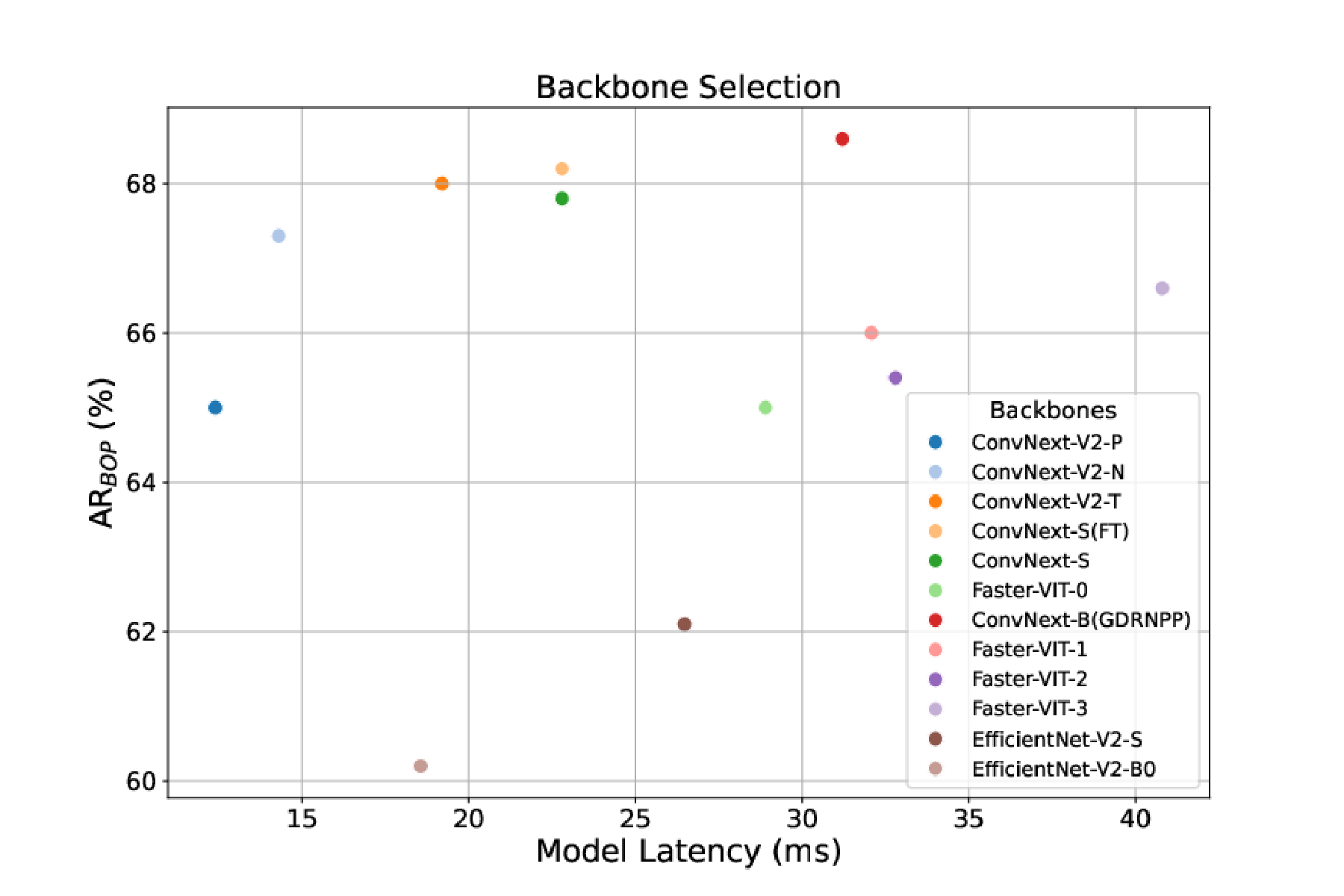}
    \caption{ Performance of various backbones when integrated into GDRNPP. }
    \label{fig:backbone}
\end{figure}

The backbones listed in Table \ref{tab:backbone} are used to estimate object poses in the LM-O dataset. All models are trained for 40 epochs with the learning rate and hyper-parameters as reported for GDRNPP. The trained models are tested on the official BOP test data, which comprises 200 images with 1445 object instances across various scenes, all eight objects being visible in almost every image. Figure \ref{fig:backbone} shows AR results for different GDRNPP backbone variants.

Despite FasterVIT variants having a much higher frame rate than ConvNext-B in ImageNet classification, their performance in pose estimation varies. The fastest variant, FasterVIT-0, is 7.39\% faster than GDRNPP but suffers a 5.3\% drop in AR. Other FasterVIT variants are slower and perform below the benchmark.
EfficientNet variants show improved speed over ConvNext-B: EfficientNet-V2-B0 is 40.53\% faster, and EfficientNet-V2-S is 15.17\% faster. However, their performance drops by 12.24\% and 9.47\%, respectively.
ConvNext variants demonstrate promising performance, especially ConvNext-V2-P: The maximum recall score drop is a minor 5.24\% while latency is reduce by 60\%. 

\textbf{Selection Criteria.} Our goal is to achieve an inference rate as close as possible to 30 frames per second or above, giving us a time budget of approximately 33.33 ms per image. With GDRNPP requiring an off-the-shelf detector like YOLOX-X (for which we measured and inference time of 8 ms), we have 25.3 ms left for pose estimation. Therefore, we focus on backbones with latencies under 25 ms, narrowing our selection to ConvNext-V2-T, ConvNext-V2-N, and ConvNext-V2-P models. ConvNext-V2-N reduces latency by about 5 ms compared to ConvNext-V2-T, with only a 1.02\% difference in AR scores. Given this trade-off, ConvNext-V2-N is chosen for its balanced performance and ConvNext-V2-P for its very low latency, allowing an extra 7-8 ms that might be invested in the optional pose refinement step. Also, keep in mind that inference time increases with the number of objects within a scene, so larger sets tend to require substantially more time. 

\subsection{Number of Regions}
\label{sec:regions}
\begin{figure}[!h]
    \centering
    \includegraphics[width=0.6\linewidth]{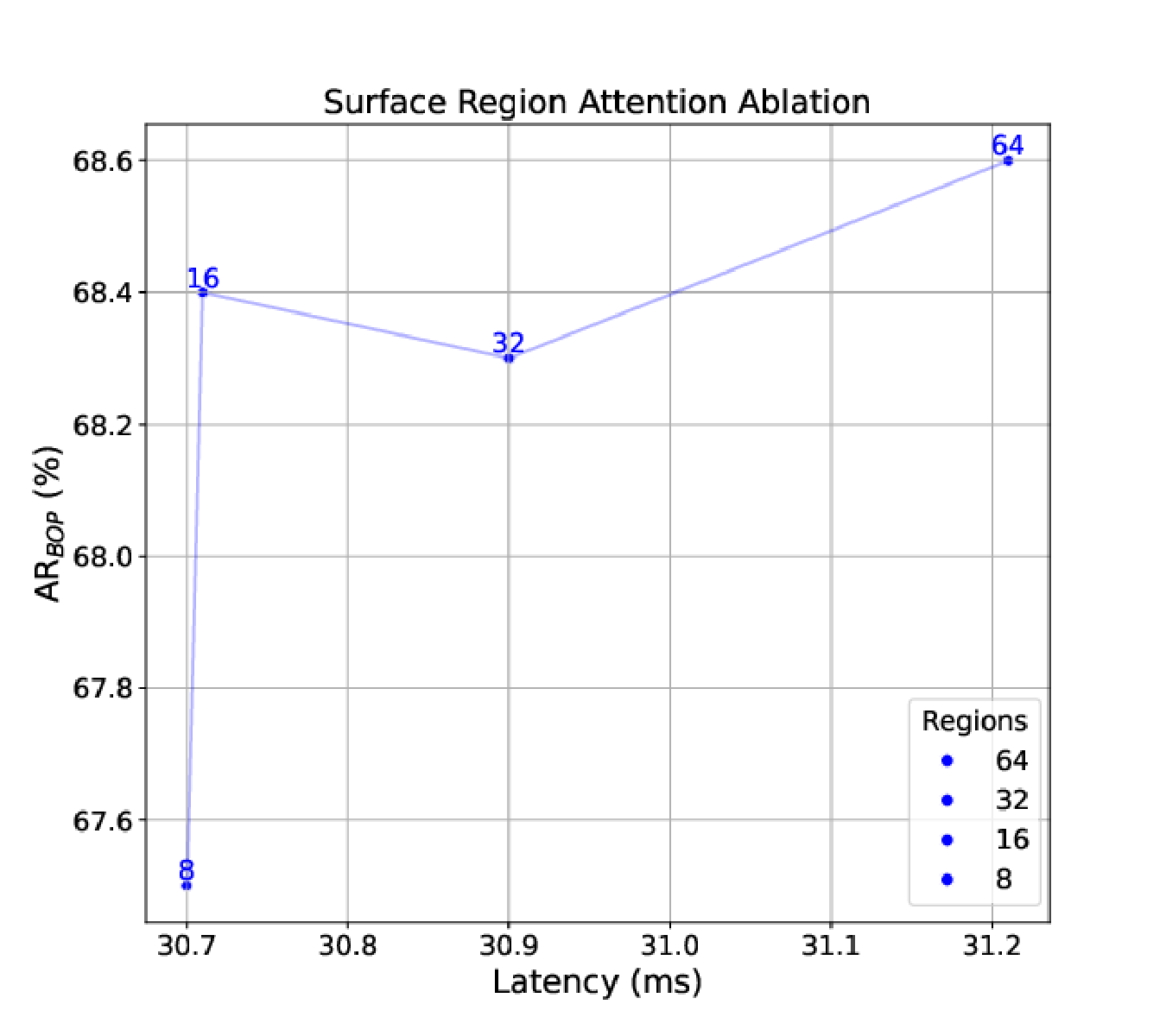}
    \caption{Illustration of performance of GDRNPP with varying number of predicted regions.}
    \label{fig:prune}
\end{figure}

As described in Section \ref{sec:attentionAblation}, we evaluate the influence of the number of regions used with surface attention on run-time. The original GDRNPP achieves an AR of 68.6\% when using 64 regions, with an inference time of 31.2ms. We test different numbers of regions and plot the results in Figure \ref{fig:prune}. As expected based on the GDR-Net paper, performance drops with lower number of regions (1.6\% with only 8 regions). 16, 32, and 64 regions offer similar AR scores, while inference speed varies by 0.5ms between the choice of 16 and 64 regions. These numbers indicate that we can only expect limited gains, when decreasing the number of regions. Considering the ablation results of original GDR-Net, which show 64 regions to be the sweet spot, we decide not to alter from their choice and also use 64 regions with our faster versions.

\subsection{Pruning}
\label{sec:EvalPruning}

\begin{figure}[h]
    \centering
    \begin{minipage}{0.5\linewidth}
    \centering
    \includegraphics[height=2.8cm]{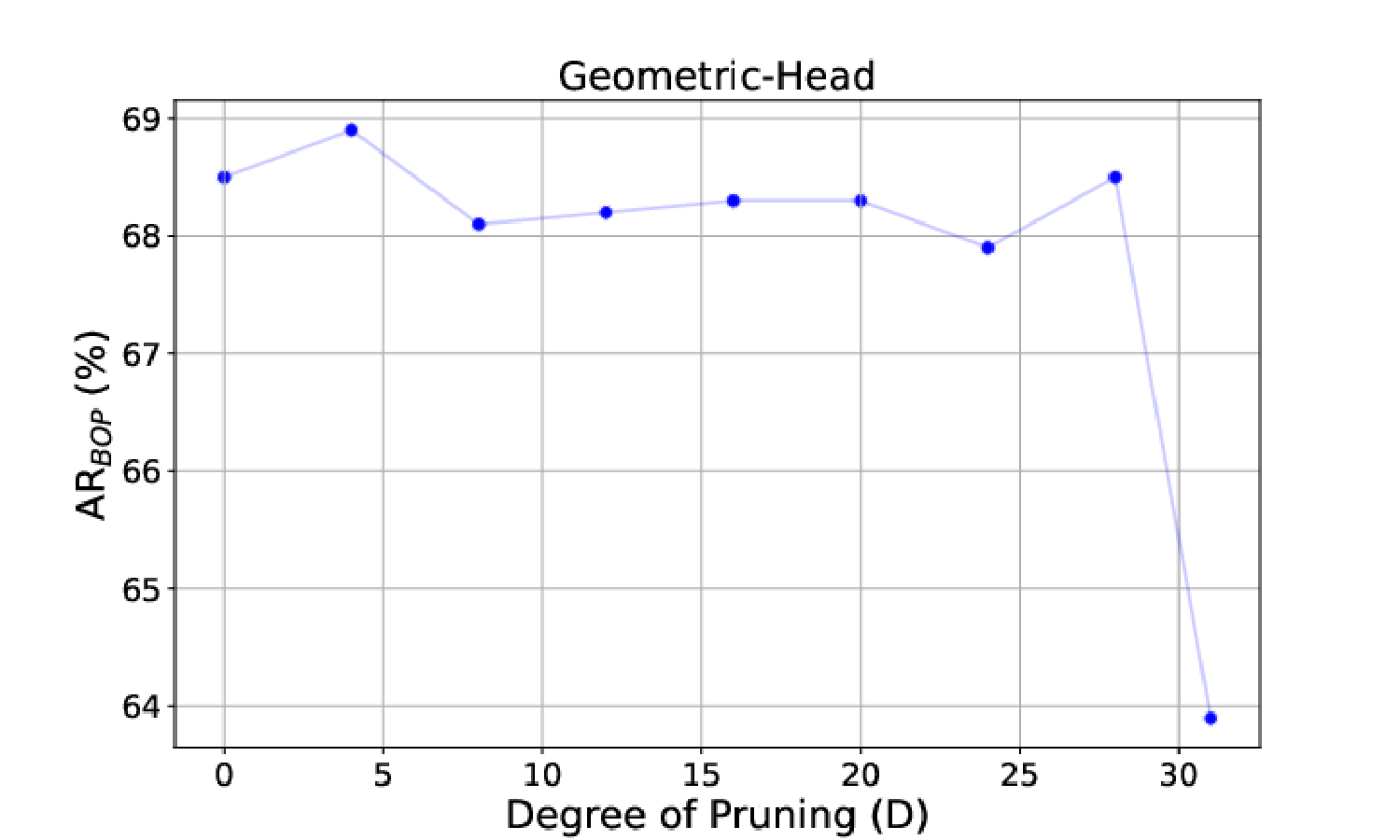}
    \caption*{(a)}
    \label{fig:avg_recall}
    \end{minipage}\hfill
    \begin{minipage}{0.5\linewidth}
    \centering
    \includegraphics[height=2.8cm]{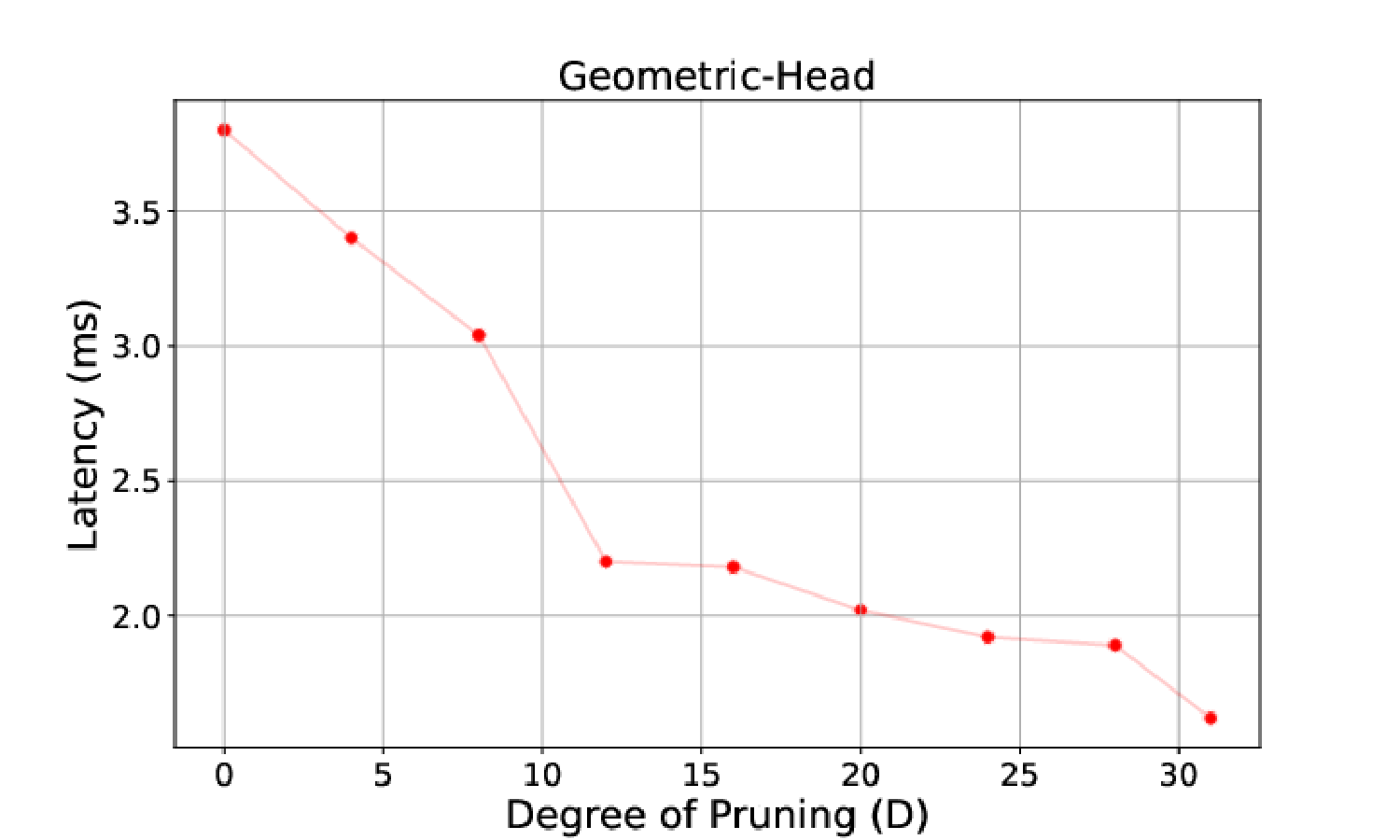}
    \caption*{(b)}
    \label{fig:latency}
    \end{minipage}\hfill
    \caption{(a) Performance of the geometric head with increasing level of pruning. (b) Latency while increasing the level of pruning.}
    \label{fig:ghead-prune}
\end{figure}

Having selected a faster backbone in Section \ref{sec:backselec}, we now focus on the geometric head and the Patch P$n$P module. The experiments described in Section \ref{sec:pruning} yield the following observations: As expected and visualized in Figure \ref{fig:ghead-prune}, increasing the degree of pruning improves latency. The original model's geometric head has a latency of 3.8ms. Initial pruning improves latency by approximately 10\%. For higher degrees of pruning, latency improves to about 1.89ms at D=28. The maximum degree of pruning in our setup is D=31, reducing latency by 57.36\%. Interestingly, AR remains stable until D=28 as long as we fine-tune the pruned model. At the maximum pruning degree, the AR drops by only 6.7\%, which is a small loss compared to the reduction in latency.

\begin{figure*}[t]
    \centering
    \includegraphics[width=\textwidth]{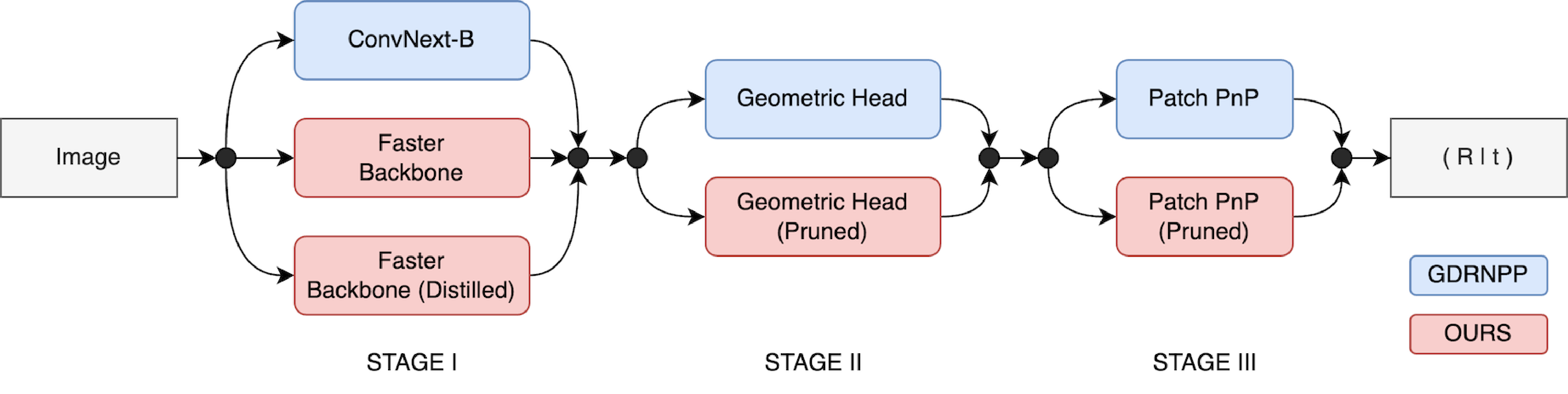}
    \caption{ Overview of our acceleration options in comparison to the baseline algorithm. }
    \label{fig:full_approach}
\end{figure*}

\begin{figure}[h]
    \centering
    \begin{minipage}{0.5\linewidth}
    \centering
    \includegraphics[height=2.8cm]{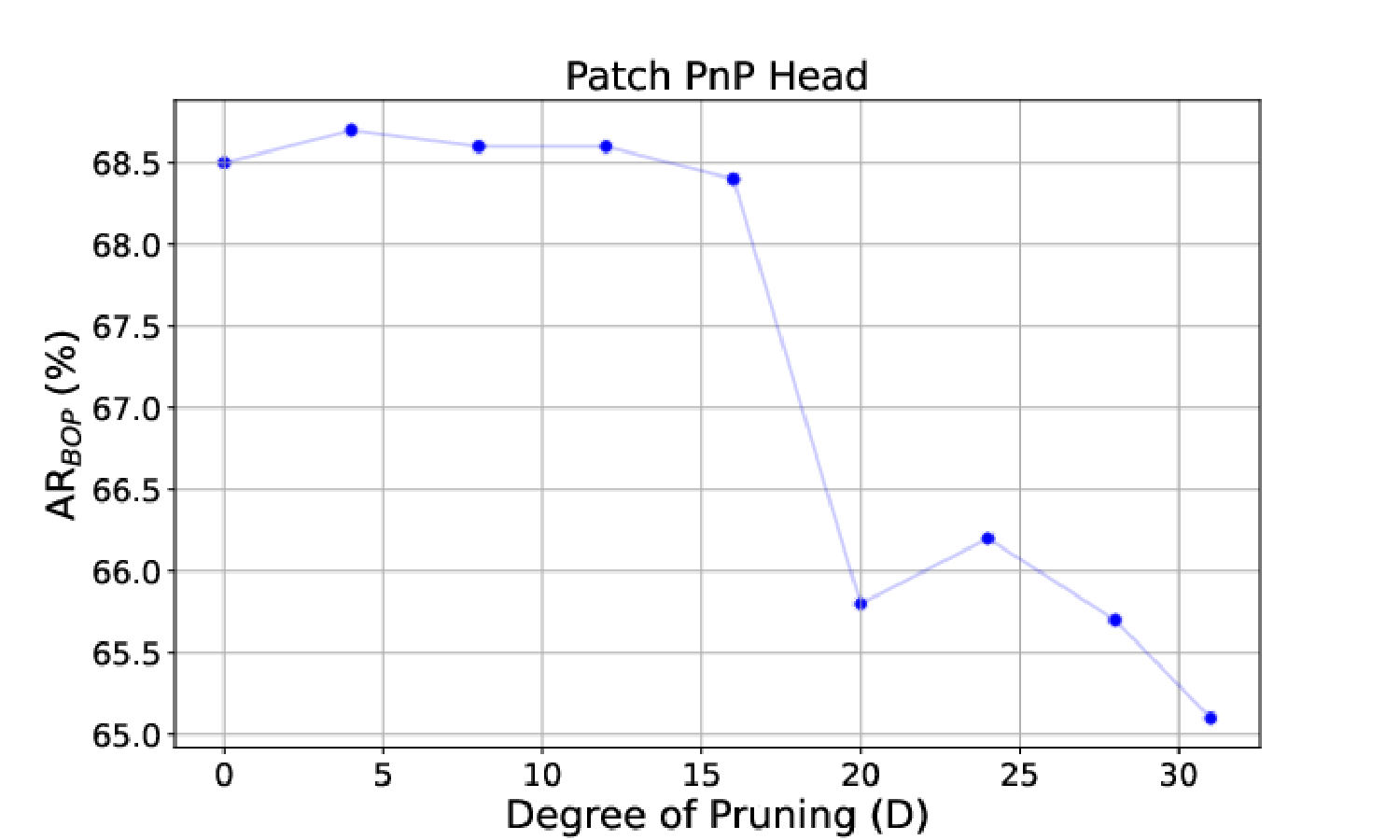}
    \caption*{(a)}
    \label{fig:pnp_avg_recall}
    \end{minipage}\hfill
    \begin{minipage}{0.5\linewidth}
    \centering
    \includegraphics[height=2.8cm]{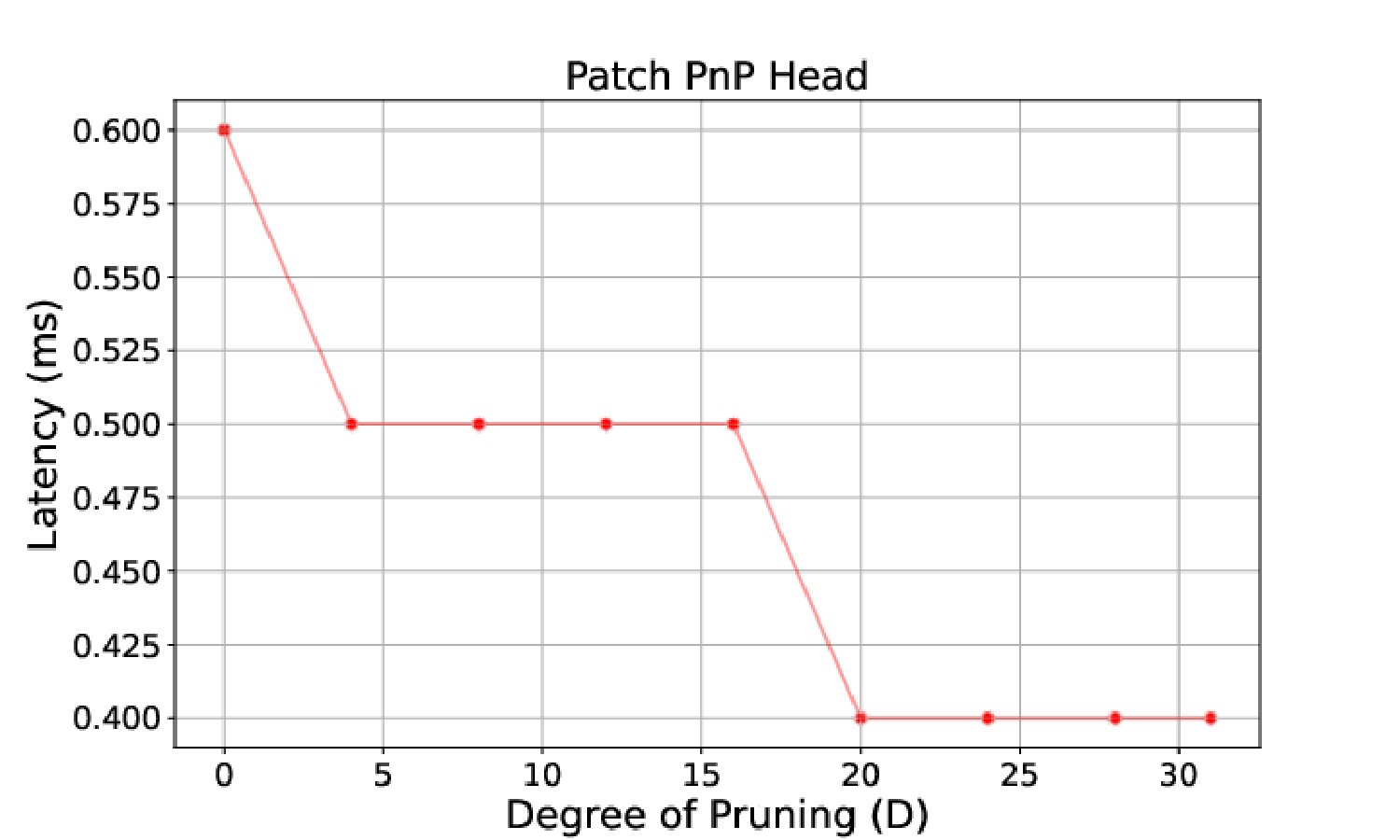}
    \caption*{(b)}
    \label{fig:pnp_latency}
    \end{minipage}\hfill
    \caption{(a) Performance of Patch P$n$P with increasing level of pruning. (b) Latency while increasing the level of pruning.}
    \label{fig:pnp-prune}
\end{figure}

Turning our attention to the the Patch P$n$P module, as illustrated in Figure \ref{fig:pnp-prune}, the unpruned Patch P$n$P module has 15.6\% lower latency than the geometric head due to fewer operations, indicating that the pruning of the Patch P$n$P module is less significant and more attention should be spent on the geometric head. Also, at higher pruning degrees (D={24, 28, 31}), latency stagnates around 0.4ms. Performance drops by 4-5\% only after pruning exceeds D=16, without significant latency improvement. Based on these observations, we decide to set our pruning hyperparameter for the geometric head to D=28 and for the Patch P$n$P to D=16, a choice that leads to no loss in performance, all the while improving latency. 

\begin{table*}[t]
\centering
\begin{tabular}{|c|c|c|c|c|c|c|c|c|c|c|}
\hline
Metrics & N & N28 & P & P28 & N-R & N28-R & P-R & P28-R & BM & BM-R \\
\hline
AR (\%) & 67.37 & 64.18 & 64.91 & 63.27 & 75.38 & 71.6 & 73.48 & 71.21 & 68.0 & 75.9 \\
\hline
Latency (ms) & 49.2 & 32.2 & 43.79 & 26.8 & 198.8 & 181.8 & 193.2 & 176.4 & 152.9 & 324.6 \\
\hline
\end{tabular}
\caption{Scores and corresponding measured latency aggregated across the seven core datasets. P28 is the fastest configuration, while N-R is the best performing.}
\label{tab:performance_aggregate}
\end{table*}

\subsection{Distillation}
\label{sec:distillation}

\begin{table}[]
\centering
\begin{small}
\begin{tabular}{|c|c|c|}
\hline
Backbone & AR wo. Dist. & AR w. Dist. \\
\hline
ConvNext-B & 68.6 & - \\
\hline
ConvNext-S & 68.2 & 69.0 \\
\hline
ConvNext-V2-T & 68.0 & 68.7 \\
\hline
ConvNext-V2-N & 67.3 & 68.0 \\
\hline
ConvNext-V2-P & 65.0 & 66.0 \\
\hline
\end{tabular}
\end{small}
\caption{Comparison of distillation performance using 4 relevant bacbkones. We see a slight outperformance when applying distillation across all experiments.}
\label{tab:distill}
\end{table}

We conduct the distillation process as described in Section \ref{sec:distillation} and compare the performance of the distilled student models to their non-distilled counterparts. The results from the experiment using different student backbones distilled from the GDRNPP model on the LM-O dataset are listed in Table \ref{tab:distill}. The four fastest backbones from Section \ref{sec:backselec} were chosen as the student backbones for this experiment. The original GDRNPP teacher model (ConvNext-B) achieves an Average Recall of 0.686. 
We see that every distilled model (w. Dist.) outperforms the model, which is trained directly on the data (wo. Dist) and it is evident that distillation helps improve the model's score. For ConvNext-S and ConvNext-V2-T, the models perform on par with the original ConvNext-B model even without distillation. However, distillation helps these models match or slightly exceed the original model's performance. Overall distillation adds a slight performance increase to the smaller backbones, at the cost of requiring a trained base model.

\subsection{Full Approach}
\label{sec:finalApproach}

\begin{figure}[!h]
    \centering
    \includegraphics[width=0.8\linewidth]{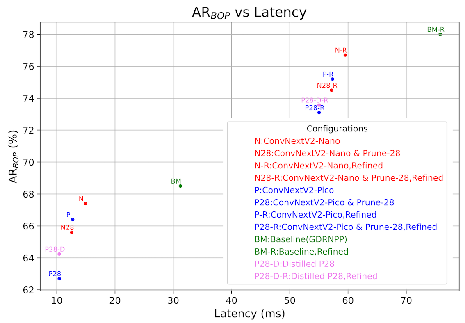}
    \caption{ Performance of our selected configurations, using distillation and/or pruning on the LM-O dataset. }
    \label{fig:lmo_with_distillation}
\end{figure}

Finally, we combine our learnings to create models for different run-time requirements. We choose two default configurations: The first opts to maximize inference speed and based on ConvNext-Pico (configuration P), the second tries to balance speed and the loss in accuracy and is based on ConvNext-Nano (configuration N). 

Based on our extensive analysis we have found the following methods to be the most promising: First, selection of a faster backbone, second backbone distillation, third pruning the geometric head and the Patch P$n$P module. We combine these into the following set of configurations: Models N and P are only replacing the backbone. 28 indicates that we use a pruned geometric head. -D shows that we use a distilled backbone. For all configurations, we additionally test the influence of fast depth refinement, indicated by -R. For comparison with the baseline, we add BM and BM-R representing the default configuration of GDRNPP. Results of the methods  on LM-O are presented in Figure \ref{fig:lmo_with_distillation}. Based on these results, we see that all models fall at different spots in the AR vs. latency ratio, giving practitioners the chance to choose what best fits their needs. 

For a more complete picture of how the methods perform on a wide range of different objects, we continue to evaluate our configurations on the remaining 6 BOP core datasets, with the omission of backbone distillation, since we found it impractical due to the training overhead. Distillation necessitates the training of an expressive teacher model, dramatically increasing overall training time and complexity. Based on the results on LM-O, we see a noticeable improvement with distillation and depending on the use case, the additional training is acceptable. However, for our remaining evaluation, we focus on faster backbones and pruning.

Figure \ref{fig:teaser} shows the average performance of our selected configurations N and P, as well as the baseline BM, with and without pruning and with and without refinement. The baseline model BM has an AR score of 68\%, which improves to 75.9\% with refinement, while the latency increases from 153ms to 324.6ms. The unpruned, unrefined variants (N and P models) show a latency improvement of 65\% and 75\% compared to the baseline, with a negligible performance drop for N and a 3\% drop for P. Pruning further improves latency, with N28 showing a 79\% improvement and P28 an 82\% improvement, accompanied by a 3.8\% and 4.7\% performance drop, respectively. Refinement improves scores for all configurations. Compared to the baseline BM-R, the unpruned models N-R and P-R have a minimal recall drop of 1.6\% and 2.4\%, respectively, with a speed improvement of 38.7\% and 40\%. Pruned versions are even faster, with N28-R and P28-R showing a recall drop of about 5\% and a latency improvement of about 44\% compared to BM-R. The fastest configuration is P28, with a 12.6\% score drop compared to BM-R, but a 92\% improvement in latency. Our best-performing configuration is N-R, with less than a 1\% score drop and a 38.7\% latency improvement. All results are shown in Table \ref{tab:performance_aggregate}.

\subsubsection{Parameter Considerations}

\begin{table}[h]
\centering
\begin{tabular}{|c|c|c|c|c|c|}
\hline
Metrics & N & N28 & P & P28 & BM \\
\hline
\# Parameters (M) & 29.4 & 24.298 & 22.68 & 17.83 & 102.87 \\
\hline
\end{tabular}
\caption{Number of parameters in millions for various configurations on the YCB-V dataset.}
\label{tab:performance_no}
\end{table}

Table \ref{tab:performance_no} shows the total number of learnable parameters for each configuration (without refinement) on the YCB-V dataset. The baseline model has 102.87M parameters. Switching to ConvNext-V2-Nano (N) and ConvNext-V2-Pico (P) reduces parameters by 71\% and 78\%, respectively. Pruning the geometric head by D=28 reduces approximately 5M parameters for both models. The pruned models N28 and P28 have a parameter reduction of 78.5\% and 82.6\%, respectively. Similar trends are observed across all datasets, with minor variations depending on the number of objects.

\subsubsection{Real-time Considerations}

\begin{table}
\centering
\small
\resizebox{\linewidth}{!}{
\begin{tabular}{|c|c|c|c|c|c|c|c|c|}
\hline
    &     & LM-O & ICBIN & YCB-V & T-LESS & ITODD & TUD-L & HB    \\
\hline
P28 & AR (\%)  & 62.7 & 59.1  & 68.4  & 68.4   & 29    & 78.5  & 76.8  \\
    & Lat (ms) & 10.4 & 60.1  & 9.8   & 35.3   & 22.7  & 8.6   & 41.1  \\
\hline
N28 & AR (\%)   & 65.6 & 59.9  & 74.2  & 68.9   & 25.4  & 80.6  & 74.7  \\
    & Lat (ms) & 12.6 & 76    & 10.8  & 37.4   & 27.9  & 9.5   & 51.5  \\
\hline
BM  & AR (\%)   & 68.5 & 63.4  & 76.8  & 77.6   & 26    & 82.8  & 80.9  \\
    & Lat (ms) & 31.2 & 425.6 & 27.7  & 167.3  & 121.7 & 13    & 284.2 \\
\hline
\end{tabular}}
\caption{AR score and inference time of our fastest configurations and the baseline reported per dataset.}
\label{tab:timings_per_set}
\end{table}

As stated above, a typical camera framerate is 30 fps, giving a total of 33.33ms of processing time per frame. Looking at our results averaged across the seven datasets, all of different complexity, we see that only P28 and N28 stay below that budget. To illustrate the effect of different datasets, we report the results of both configurations per dataset in Table \ref{tab:timings_per_set}. We see vast differences in inference speed per dataset. Larger datasets such as T-LESS, ITODD, HB, and ICBIN tend to have more objects per image and much higher compute requirements. At the same time, our configurations move much closer towards real-time across all datasets, fullfilling the 30 fps target on LM-O, YCB-V, ITODD, and TUD-L. Our fastest configuration achieves 16.6 fps on the slowest dataset (ICBIN) compared to 2.35 ms with the original algorithm.

\section{Conclusions and Future Work}
\label{sec:conclusion}
We explored acceleration techniques for the 6D pose estimator GDRNPP. Surface region attention improved performance but not inference time, while smaller backbones enhanced speed with minimal performance loss. Knowledge distillation offered slight gains but requires the training of 2 individual models. The geometric head has a greater impact on inference time than the Patch P$n$P modules, leading us to reduce its parameters via pruning. We developed configurations that balance accuracy and speed, notably configuration P28, which has a 4.7\% performance drop and an 82\% latency improvement. Our findings move state-of-the-art 6D pose estimation a step closer towards real-time inference.

\bibliographystyle{eg-alpha-doi} 
\bibliography{egbibsample}       


\end{document}